\documentclass[10pt,twocolumn,letterpaper]{article}

\usepackage{cvpr}              
\usepackage{graphicx}
\usepackage{amsmath}
\usepackage{amssymb}
\usepackage{booktabs}
\usepackage{times}
\usepackage{helvet}
\usepackage{url}
\usepackage{courier}
\usepackage{tabularx}
\usepackage{times}
\usepackage{epsfig}
\usepackage{diagbox}
\usepackage{makecell}
\usepackage{graphicx}
\usepackage{caption}
\usepackage{float}
\usepackage{amssymb}
\usepackage[none]{hyphenat}
\definecolor{cvprblue}{rgb}{0.21,0.49,0.74}
\usepackage[pagebackref,breaklinks,colorlinks,allcolors=cvprblue]{hyperref}
\usepackage{multirow}
\def\paperID{1} 
\def\confName{CVPR}
\def\confYear{2025}

\title{CACP: Context-Aware Copy-Paste to Enrich Image Content for Data Augmentation}

\author{%
  Qiushi Guo$^{1}$\ \ \ 
  Shaoxiang Wang$^{2}$\ \ \ 
  Chun-Peng Chang$^{2}$\ \ \ 
  Jason Rambach$^{2}$\thanks{Corresponding author}\\
    $^{1}$CSRD, Chengdu, China\\
    $^{2}$German Research Center for Artificial Intelligence (DFKI), Kaiserslautern, Germany\\
  \texttt{guoqiushi910@gmail.com}\\
  \texttt{\{shaoxiang.wang,chun-peng.chang,Jason.Rambach\}@dfki.de}
}

\begin{document}
\maketitle
\begin{abstract}
   Data augmentation is a widely used technique in deep learning, encompassing both pixel-level and object-level manipulations of images. Among these techniques, Copy-Paste stands out as a simple yet effective method. However, current Copy-Paste approaches either overlook the contextual relevance between source and target images, leading to inconsistencies in the generated outputs, or heavily depend on manual annotations, which limits their scalability for large-scale automated image generation. To address these limitations, we propose a context-aware approach that integrates Bidirectional Latent Information Propagation (BLIP) for extracting content from source images. By aligning the extracted content with category information, our method ensures coherent integration of target objects through the use of the Segment Anything Model (SAM) and YOLO. This approach eliminates the need for manual annotation, offering an automated and user-friendly solution. Experimental evaluations across various datasets and tasks demonstrate the effectiveness of our method in enhancing data diversity and generating high-quality pseudo-images for a wide range of computer vision applications.
\end{abstract}
\section{Introduction}
Deep Learning-based approaches have become the major paradigm in many computer vision tasks, ranging from classification to segmentation. These approaches outperform traditional ones in terms of accuracy and generalization. However, the bottleneck of supervised deep learning is the quality and quantity of the training dataset. To obtain a 
dataset, a large volume of images needs to be annotated, which is labour-intensive and time-consuming. For segmentation task, annotating a single image is estimated to take up to 1.5 hours\cite{cordts2016cityscapes}. How to generate high-quality, highly realistic datasets has become an important research question in recent years. 
\begin{figure}[htb]
    \centering
    \includegraphics[width=0.5\textwidth]{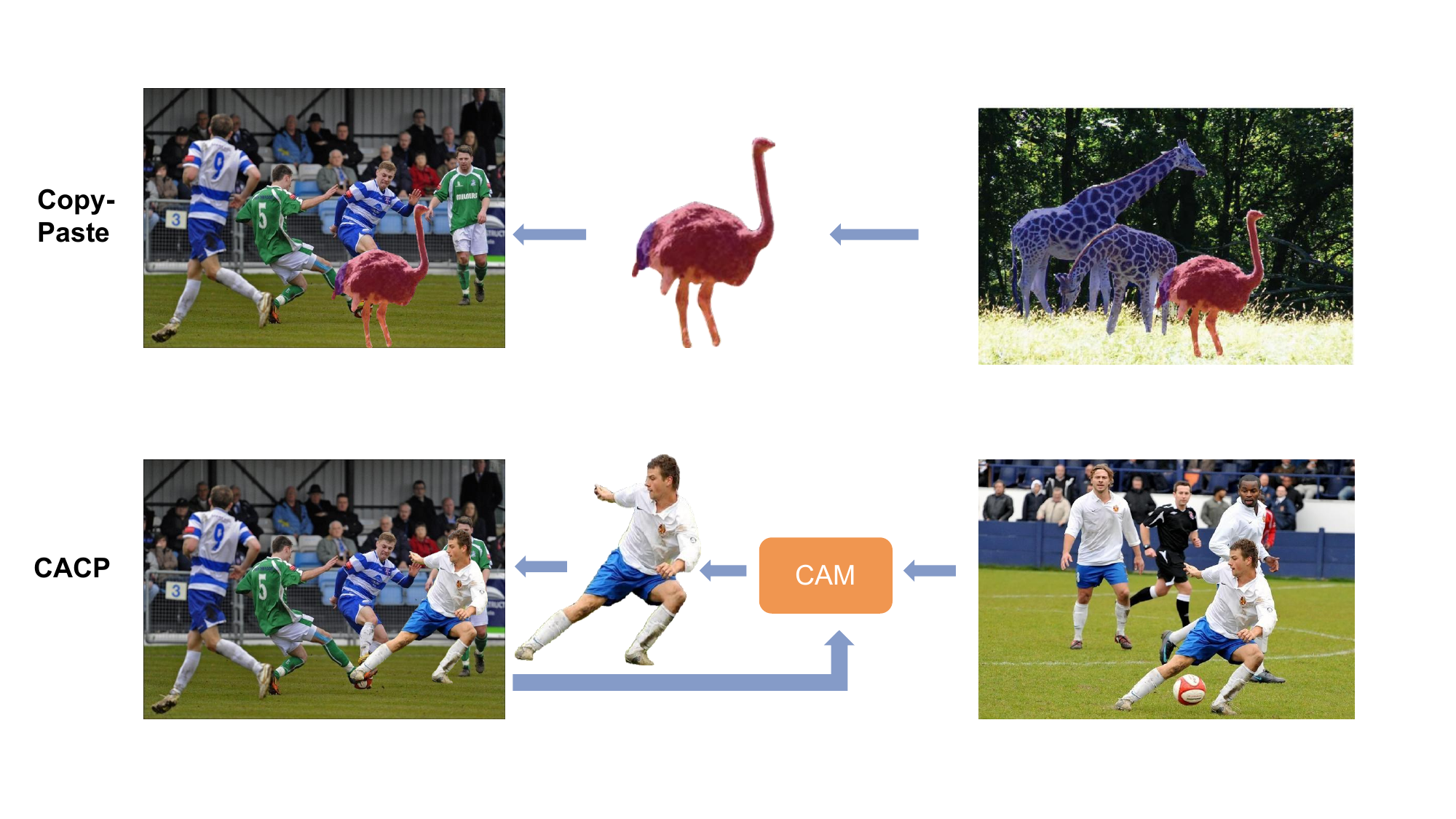}
    \caption{Comparison between the Copy-Paste method (first row) and CACP (second row). The former overlooks the contextual relevance between the base and target images, leading to disharmony. Our approach leverage the semantic information using CAM(Context-Aware-Module) to alleviate this issue. }
    \label{drawback}
\end{figure}
Previous data augmentation methods increase the diversity of images by applying operations such as flipping, rotating, and adding blur and noise. However, these techniques fail to enhance the content of images at the object level. To address this issue, the Copy-Paste method \cite{copy-paste2021simple} was proposed. The idea is straightforward and intuitive: objects from target images are pasted onto source images at random positions, resulting in images with enriched content.

However, existing Copy-Paste methods have several drawbacks:
\begin{itemize}
    \item \textbf{Context Neglect}: Methods often neglect the contextual relationship between the copied objects and the target images. For example, a penguin is unlikely to appear in a desert, and a giraffe is improbable on a soccer field. Such contextually incompatible images reduce the practical significance of the augmented dataset.
    \item \textbf{Dependency on Masks}: The original Copy-Paste approach depends on publicly available image-mask pairs to generate new images, which limits its applicability and requires additional effort to extend its use. This process is not adaptable to scenarios where masks are unavailable.
\end{itemize}

To address the aforementioned issues, we propose a novel approach named Context-Aware Copy-Paste (CACP), leveraging large language models (LLMs) and vision foundation modles. Our method integrates several NLP-based models to ensure contextual relevance between the source and target images. The main procedure is as follows:
A vision-language model is used to generate captions for the source images (the images to be augmented). Object365, a dataset containing 365 distinct classes, serves as the target image set. For a given source image, a similarity score is computed between its caption and the category names in Object365 using a semantic similarity model based on a transformer architecture. The category with the highest similarity score is selected, and an image from this category is randomly chosen as the target image. An object detection model is then employed to identify objects in the target image. The bounding box of the detected object is processed by a segmentation model to obtain a pixel-level mask. Finally, the object, guided by the mask, is pasted onto the source image.

Our approach can be applied to several computer vision tasks, including classification, object detection, and segmentation. Without requiring extra manual annotation, the target gallery can be easily extended to adapt to specific tasks. 
Our contributions can be summarized as follows:
\begin{itemize}
    \item We propose a data augmentation mechanism called Context-Aware Copy-Paste (CACP), which semantically bridges the source and target images. Additionally, this approach is easily extendable to custom tasks without requiring extra annotation.
    \item We demonstrate that robust segmentation results can be achieved by combining object detector and class activation mapping as prompt generators.
    \item The experiments results demonstrate that our method outperforms the original Copy-Paste technique and enhances model performance.
\end{itemize}


\begin{figure*}[htb]
    \centering
    \includegraphics[width=0.98\textwidth]{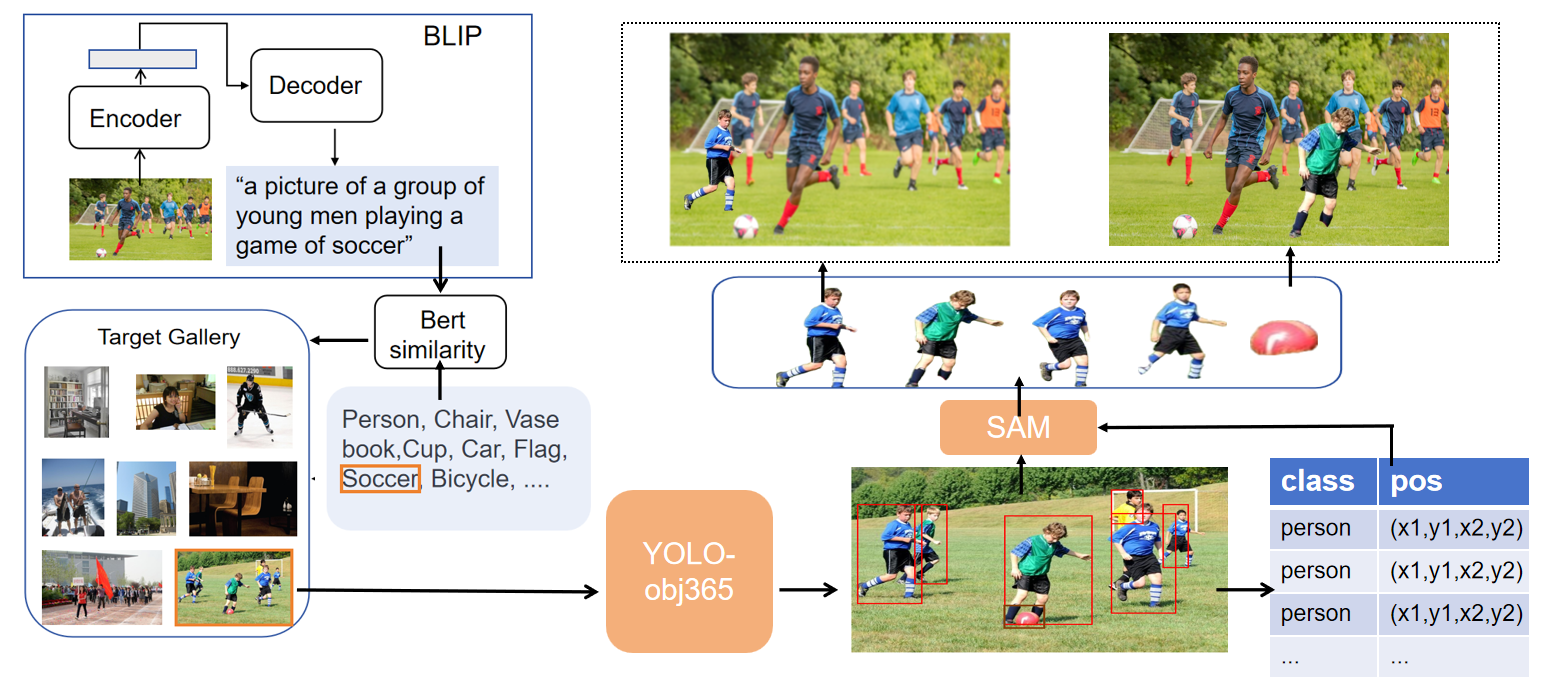}
    \caption{Our method's pipeline involves leveraging BLIP and BERT to select the best-matched target image from a gallery. Subsequently, the corresponding mask is obtained using YOLO and SAM. A single base-target pair can generate multiple augmented images based on user preferences.}
    \label{workflow}
\end{figure*}
\section{Related Work}
\subsection*{Copy-Paste for Data Augmentation}
Copy-Paste has been widely used in semi-supervised scenarios due to its efficiency \cite{guo2023real,guo2024universal}. Dvornik et al.\cite{dvornik2018modeling} first proposed the copy-paste approach for object detection tasks based on visual context, which significantly boosted performance on the VOC07 \cite{pascal-voc-2007} dataset. However, they only used VOC2012\cite{pascal-voc-2007} as their target gallery, making it challenging to apply the method to other specific scenarios. Additionally, they used a CNN classifier to obtain context information(describing an image only by a word), which is less effective compared to our BLIP-based approach(describing an image by a sentence). Golnaz et al.\cite{copy-paste2021simple} was the first to propose the Copy-Paste data augmentation method for instance segmentation.
They claim that simply pasting objects randomly provides substantial gains over baselines. Although their approach is easy to implement, the random pasting generates images that lack the grounding of real images, as the distribution of object co-occurrences is ignored. Zhao et al. \cite{x-paste2023x} proposed X-Paste, which leverages zero-shot recognition models like CLIP to make the approach scalable. X-Paste demonstrated impressive improvements over CenterNet2. Viktor Olsson el al.\cite{olsson2021classmix} introduced ClassMix, which generates augmentations by mixing unlabeled samples based on the network's predictions to respect object boundaries. Inspired by their work, we combined vision-language models with copy-paste to generate augmented images efficiently.
\subsection*{Vision Language Models and Zero-shot Segmentation}



Vision-Language Pre-training (VLP) has recently made significant breakthroughs. The zero-shot capability and image-text alignment make it an ideal support for the copy-paste pipeline.

CLIP \cite{clip2021learning} is a neural network trained on a variety of (image-text) pairs. It can be instructed in natural language to predict the most relevant text snippet for a given image, without direct task-specific optimization. MaskCLIP\cite{dong2023maskclip}incorporates a newly proposed masked self-distillation into contrastive language-image pretraining, it distills representation from a full image to the representation predicted
from a masked image. BLIP \cite{li2022blip} is a new VLP framework that flexibly transfers to both vision-language understanding and generation tasks. Utilizing noisy web data, BLIP achieves state-of-the-art results on several benchmarks and performs exceptionally well on zero-shot tasks.

SAM is a promptable instance segmentation model trained on the largest segmentation dataset to date \cite{kirillov2023segment}. It can generalize to new, unseen distributions and tasks, with competitive or even superior performance compared to prior fully supervised results. However, SAM's performance depends heavily on the quality of prompts; insufficient prompts can lead to unstable or unintended segmentation results. SAM is widely used to guide data augmentation. For example, \cite{point-samaug} introduced SAMAug, a novel visual point augmentation method for SAM that enhances interactive image segmentation performance. The above approach could be improved by introducing semantic tool to bridge the source and target objects.
\section{Motivation}
Although copy-paste data augmentation methods have significantly improved computer vision tasks\cite{copy-paste2021simple}, traditional crop-paste pipelines have two notable limitations. First, the crop-paste method is challenging to scale effectively. Second, the semantic gap between the source image and the target image remains unaddressed.
\subsection{Scalability and Extendability}
Previous copy-paste methods heavily relied on image-mask pairs to perform operations. However, preparing masks for images is costly and time-consuming, making it challenging to apply these methods in a generic manner. To address this issue, previous copy-paste approaches have resorted to using public datasets with pixel-level annotations such as VOC2007 and CamVid\cite{camVid2009semantic,pascal-voc-2007}.

However, these datasets are limited in the number of categories they cover, thereby restricting the diversity of content in generated images. \cref{segmentation dataset} provides a listing of properties of several public segmentation datasets (ADE20K \cite{ade20k}, COCO \cite{coco}, VOC2007 \cite{pascal-voc-2007}). These datasets often cannot meet the specific requirements of scenarios. For instance, in a foreign object detection task\cite{shao2019objects365}, foreign objects may span hundreds of categories, making it impractical to rely on public datasets or manual annotation to prepare the dataset. An entirely automated copy-paste pipeline is needed to generate large quantities of high-quality images.
\begin{table}[h]
    \centering
    \begin{tabular}{ccccc}
    \toprule
    dataset  & images& classes &resolution \\
    \midrule
        VOC2007\cite{pascal-voc-2007}& 9963 & 20&-\\
        CamVid\cite{camVid}&701 & 32&480*360&\\
        CityScape\cite{cityscapes}&20000&30&2024*1048&\\
        coco\cite{coco}&330k &80 &640*480&\\
        ADE20K\cite{ade20k}&25574  &150&1650*2220&\\
    \bottomrule
    \end{tabular}
    \caption{Summary of properties of common public segmentation datasets}
    \label{segmentation dataset}
\end{table}
\subsection{Content Discrepancy}

\begin{figure*}[htb]
    \centering
    \includegraphics[width=0.98\textwidth]{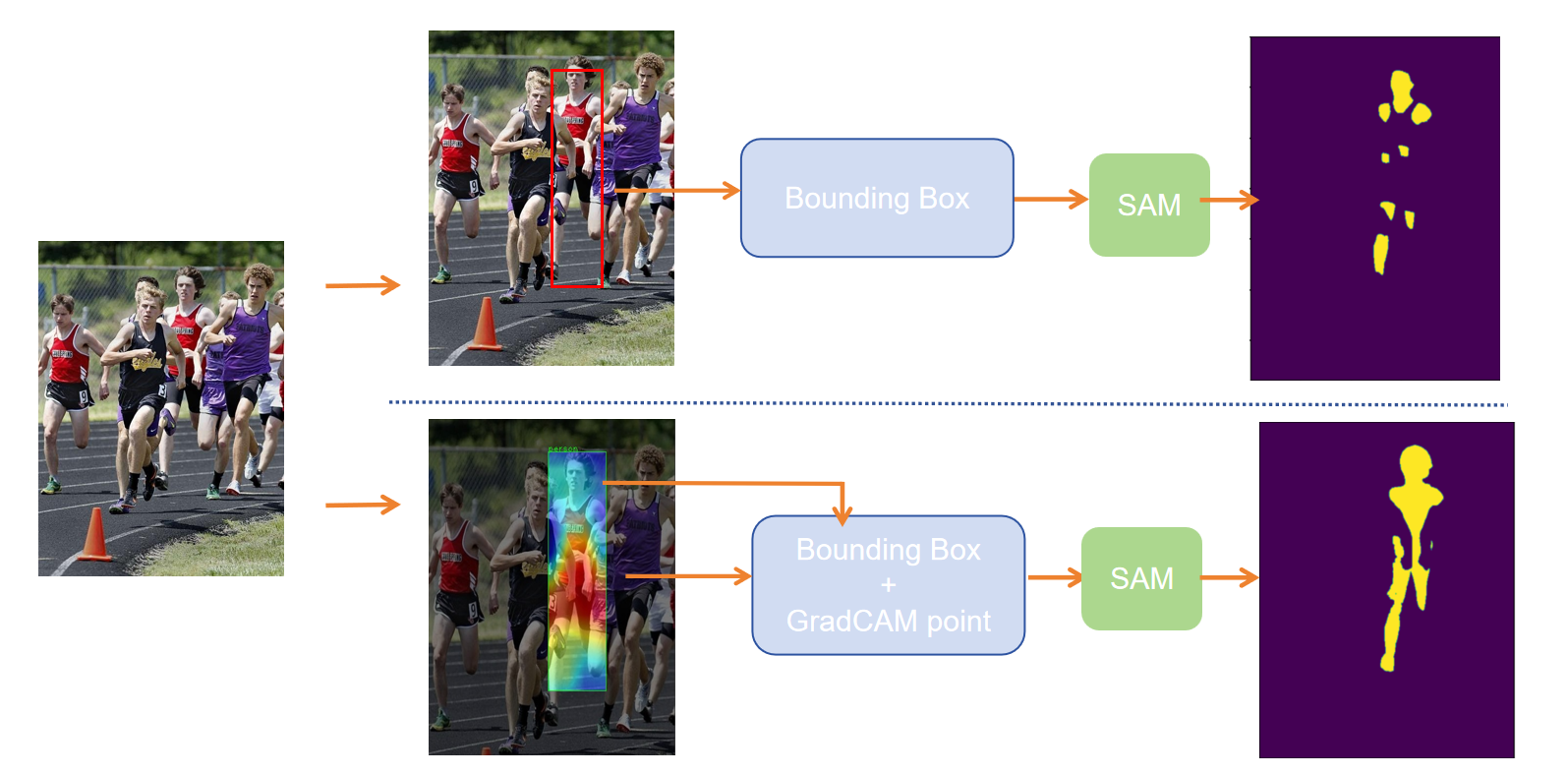}
    \caption{Comparison of SAM segmentation results using different prompts: Single bounding box prompts (upper row) tend to produce incomplete masks, while combining bounding boxes with Grad-CAM points generates more accurate and robust masks.}
    \label{cam-sam}
\end{figure*}

Previous work has often overlooked the issue of content relevance in the data augmentation process. Irrelevant objects are frequently pasted onto source images, providing minimal training benefit. As depicted in \cref{cam-sam} , images generated by traditional copy-paste methods contribute less to person-related computer vision tasks, as evidenced by corresponding Grad-CAM results. In addition, models trained on such images failed to learn scene context, i.e. which classes are likely to co-exist in an image.

We propose that incorporating highly relevant objects into source images at appropriate positions and scales can enrich the image content and expedite the training process. Experimental results validate our hypothesis: traditionally pasted elements often fail to activate appropriately. In contrast, our method successfully triggers activations, as demonstrated by GradCam visualization techniques \cref{drawback}, enhancing the content relevant to the desired classes.


\begin{figure}[h]
    \centering
    \includegraphics[width=0.48\textwidth]{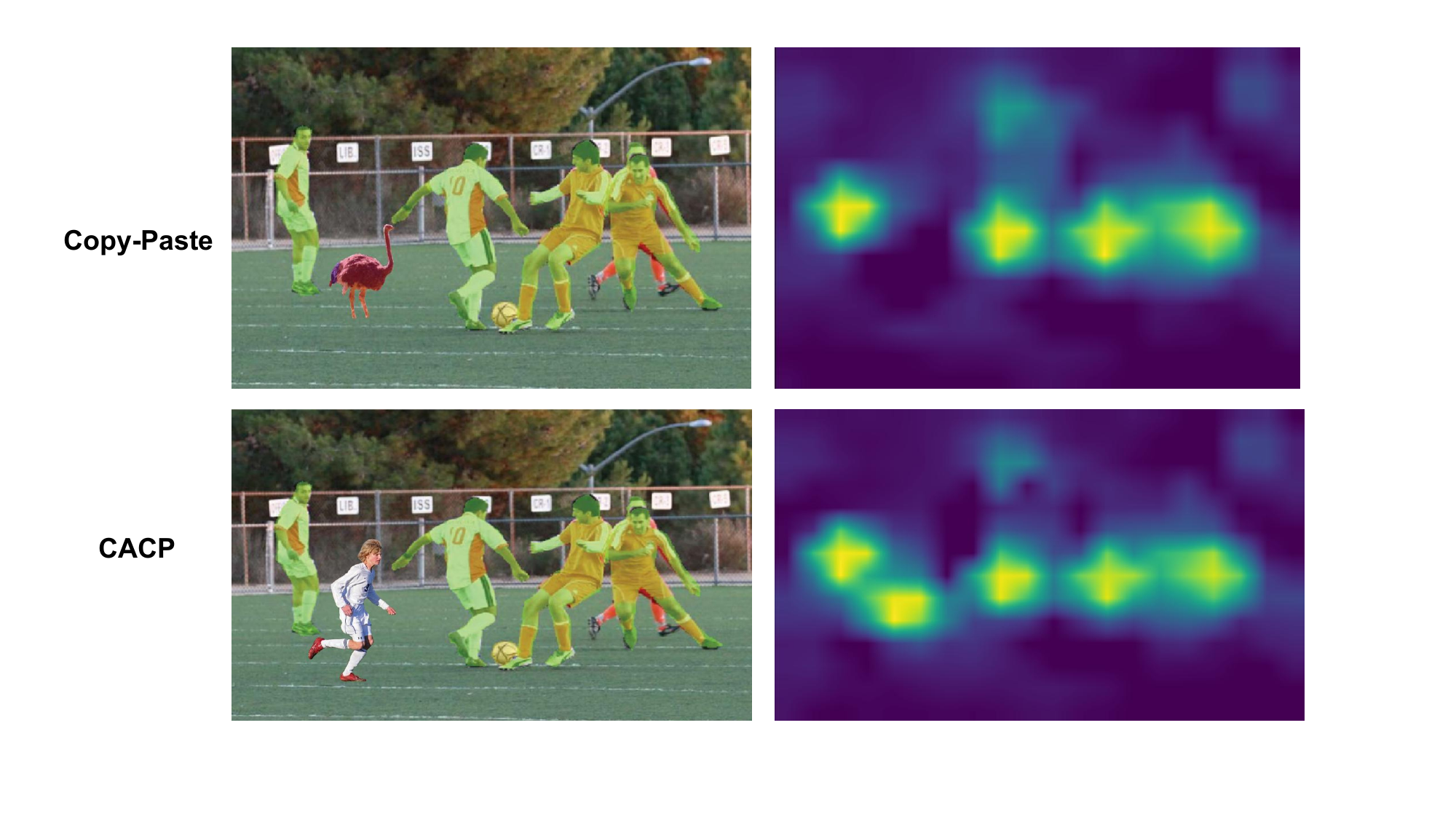}
    \caption{GradCam comparison between the Copy-Paste(top row) and our context-aware copy paste(bottom row). CACP contributes more in person related vision tasks compared to copy-paste.}
    \label{drawback}
\end{figure}
\section{Method}
Our CACP approach can be split into the following
parts: Gallery Preparation, Context-Awareness Module and Copy-Paste. In the Gallery preparation part, target images are selected to provide object-level content enhancement; In the context aware stage, source image and target image are bridged using a BLIP and BERT-based similarity measurement tool. Once the preferred target image category is determined, object of interests in target image will be cropped 
and pasted onto the source image considering the size and position. An object detector model and SAM will be leveraged to make the crop-paste process fully automatic. The details of each procedure will be described in follow paragraphs.

\subsection{Problem Formulation}
Given a source image set $\mathcal{D}_{S}$ and a target image $I_{T}$, our task is to find the most relevant image $I_{S} \in \mathcal{D}_{S}$ and most relevant object $obj\in \mathcal{O}$, where $\mathcal{O}$ is the collection of objects in $I_{S}$. Specifically, $I_{S}$ can be obtained as follow:
\begin{equation}
    I_{s} = \mathop{\arg\min\hspace{0.5em}}\limits_{I_{i}}\phi(I_{i},I_{T})
\end{equation}
where $\phi(\cdot)$ is the function to measure the semantic similarity between two images.
Once $I_{S}$ is determined, $obj$ and corresponding mask $M$ can be inferred as follow:
\begin{equation}
    obj, M = \psi(I_{S})
\end{equation}
where $\psi(\cdot)$ are deep learning models, which take images as input and output coordinates(detection task) or labels of each pixel(segmentation task).
\begin{equation}
    I_{syn} = I_{S}\otimes M +I_{T}\otimes(1-M)
\end{equation}
where $I_{syn}$ is the generated image, $\otimes$ is pixel-wise multiplication.
\subsection{Data preparation}
To enhance the diversity of our dataset, a substantial collection of images is essential for our gallery. In this study, we utilize Object365 \cite{shao2019objects365} as our image gallery. Object365 encompasses 365 classes, featuring over 2 million images and 30 million bounding boxes. These images are characterized by high resolution and quality annotations. In contrast, COCO offers only 80 classes. Object365 significantly expands the range of target objects available for augmentation.

Additionally, we propose an alternative method to leverage images without bounding box annotations, thereby enhancing the applicability of our approach. Specifically, we assume that all images in the galleries are presented without bounding boxes or masks, and each image is labeled solely with its category name. This approach enables users to extend custom categories and adapt them to specific scenarios.
\subsection{Context Awareness Module}
\subsubsection*{Image Captioning}
To establish semantic coherence between the source and target images, it is crucial to recognize the contents of both images beforehand. Rather than solely detecting objects within the images, we employ a state-of-the-art Visual-Language pre-training (VLM) model as the content extractor. In this role, we utilize BLIP\cite{li2022blip}. Compared to object-detection methods, BLIP generates smooth and natural descriptions of input images, rather than isolated words. Furthermore, while object-level approaches may struggle to provide meaningful information when encountering unseen objects, BLIP consistently offers general information applicable to common scenarios.

\subsubsection*{Target object matching}
In the last step we obtain the caption of the source image,namely $C(I_{S})$. Due to the large amount of the target image gallery, as a trade-off, we take the class name as the caption of the target image, annotated as $C(I_{T_{i}}(i=1,2,...,n))$,where n is the total number of categories. To determine the correlation between the $C(I_{S})$ and $C(I_{T_{i}})$, Bert-embedding\cite{devlin2018bert} is leveraged as our measurement tool. In Table \ref{similarity-embedding} we present examples of samples using Bert-embedding to calculate the similarity in our work compared to traditional approaches.
\begin{table}[h]
\centering
\begin{tabular}{cccc}

\toprule[1pt]
 Image caption&category &Bert  \\
\hline
\multirow{2}{*}{\makecell[c]{\textit{“Two teams are playing}\\ \textit{football games"}}}&soccer&0.94\\
&pig&0.41\\

\hline
\multirow{2}{*}{\makecell[c]{\textit{"A boy is dancing}\\ \textit{with a girl in the garden"}}}&person&0.89\\
&goose&0.46\\

\hline
\multirow{2}{*}{\makecell[c]{\textit{"A boy is standing }\\ \textit{near a red car"}}}&flower&0.51\\
&truck&0.89\\
\bottomrule[1pt]
\end{tabular}

\caption{Comparison between BERT-similarity and Cosine-similarity.}
\label{similarity-embedding}
\end{table}
From the \cref{similarity-embedding} we can find that Bert-based distance metric is preferred.

\subsection{Copy-Paste}

Once the category with the highest similarity score is determined, we randomly select an image from this class as the target image. The SAM is then employed as the pixel-level mask extractor. SAM is a single-shot segmentation model capable of segmenting any object based on prompts, such as bounding boxes or multiple points indicating the intended objects.
\subsubsection{Prompt generation}
To obtain the prompts for segmentation, we utilize YOLO-365, a model trained on the Object365 dataset, to detect objects within the target images. For instance, if the YOLO-365 model detects a 'dog' within a target image categorized under dogs, the bounding box of the dog is then forwarded to the SAM model along with the target image. SAM subsequently generates the corresponding segmentation mask for the detected object. Finally, guided by this generated mask, the pixels representing the dog are pasted onto the source images.


In our experimental trials, we observed that feeding pure bounding boxes into the Segment Anything Model (SAM) often results in unintended or incomplete masks for the corresponding target objects. To mitigate this issue, we propose an approach based on Grad-CAM \cite{gradCAM} to achieve more accurate segmentation masks.

By inputting the target image into the Grad-CAM module, the resulting heatmap provides valuable positional information about the target. We then use this heatmap to sample points, which are combined with the bounding box as prompts for the SAM model. This hybrid approach, illustrated in Figure \ref{cam-sam}, improves the accuracy of segmentation results compared to using bounding boxes alone.

\subsubsection{Scale and Position}
To enhance the realism and harmony of the generated image, rescaling and rendering techniques are implemented. The cropped objects are rescaled according to a ratio interval based on our statistical analysis of the Object365 dataset. We traverse the images in Objects365, and record ratios of image pair, namely $obj1-obj2-ratio$. Before pasting the object onto the source images, we extract object pair names and obtain the $ratio_{max}^{obj1,obj2}$ and $ratio_{min}^{obj1,obj2}$ from record. 
\section{Experiments}
\begin{table*}[t]
    \centering
    \resizebox{0.95\textwidth}{!}{
    \begin{tabular}{cccccccccc}
    \toprule
         \multirow{2}{*}{Methods}&\multicolumn{6}{c}{\textbf{CamVid}}\\
         &Car &Pedestrian&Building&Road&Sky&Tree&$\uparrow$\\
    \midrule
    \midrule
     U-Net\cite{unet}&0.776 &0.447&0.764&0.872&0.872&0.834&\\
     U-Net+CACP &0.789(+\textcolor{green}{0.013}) &0.481(+\textcolor{green}{0.034})&0.783+\textcolor{green}{0.019}&0.893(+\textcolor{green}{0.021})&0.875(+\textcolor{green}{0.003})&0.841(+\textcolor{green}{0.007})&+\textcolor{green}{0.016}\\
     \hline
     FPN\cite{fpn}&0.792 &0.432&0.783&0.897&0.884&0.867\\
     FPN+CACP & 0.813(+\textcolor{green}{0.011})&0.479(+\textcolor{green}{0.047})&0.797(+\textcolor{green}{0.014})&0.903(+\textcolor{green}{0.006})&0.885(+\textcolor{green}{0.001})&0.881(+\textcolor{green}{0.014})&+\textcolor{green}{0.015}\\
     \hline
     PSPNet\cite{pspnet} &0.788 &0.445&0.792&0.886&0.875&0.843\\
     PSPNet+CACP &0.803(+\textcolor{green}{0.015}) &0.487(+\textcolor{green}{0.013})&0.799(+\textcolor{green}{0.007})&0.901(+\textcolor{green}{0.015})&0.873(+\textcolor{green}{0.002})&0.862(+\textcolor{green}{0.019})&+\textcolor{green}{0.012}\\
     \hline
     DeepLabV3\cite{deeplabv3} &0.792&0.461 &0.803&0.908&0.891&0.875\\
     DeepLabV3+CACP &0.811(+\textcolor{green}{0.019})&0.493(+\textcolor{green}{0.028}) &0.812(+\textcolor{green}{0.009})&0.922(+\textcolor{green}{0.014})&0.887(-\textcolor{red}{0.004})&0.89(+\textcolor{green}{0.019})&+\textcolor{green}{0.011}\\
     \hline
     DeepLabv3plus\cite{deeplabv3plus}&0.803&0.471 &0.817&0.927&0.907&0.871\\
     DeepLabv3plus+CACP&0.817(+\textcolor{green}{0.014})&0.497(+\textcolor{green}{0.026}) &0.826(+\textcolor{green}{0.009})&0.933(+\textcolor{green}{0.006})&0.912(+\textcolor{green}{0.005})&0.883(+\textcolor{green}{0.012})&+\textcolor{green}{0.012}\\
     \hline
     PAN\cite{PAN}&0.794& 0.501&0.806&0.931&0.883&0.868\\
     PAN+CACP&0.812(+\textcolor{green}{0.008}) &0.513(+\textcolor{green}{0.012}) &0.821(+\textcolor{green}{0.015}) &0.937(+\textcolor{green}{0.006})&0.891(+\textcolor{green}{0.008})&0.891(+\textcolor{green}{0.023})&+\textcolor{green}{0.012}\\
           
    \bottomrule
    \end{tabular}}
    
    \caption{CACP provides robust gains across popular segmentation architectures in CamVid except \textbf{Sky} in DeeplabV3.}
    \label{across architectures}
\end{table*}
\subsection{Configuration}
The experiments are conducted in the pytorch platform. GPU is RTX 3090ti with 24GB memory.
For the classification task, the batch size is set to 16 and the loss function is cross entropy. Adam\cite{adam} is used as the optimizer with learning rate of 0.001 over 50 epochs. For the segmentation task, batch size is set to 8, loss function is dice loss,epochs are set to 20. For the detection task, we set batch size as 8, and epochs are set to 50. For segmentation task and object detection task we use the segmentation pytorch model and yolov5-s\cite{ultralytics2021yolov5}, respectively. Our segmentation models are implemented by Segmentation-Models-Pytorch(SMP) library\cite{smp}. 
\subsection{Metric}
Precision is selected as the metric for classification task, which is calculated as given below:
\begin{equation}
    Accuracy = \frac{Number\_correct\_predictions}{Total\_number\_of\_Predictions}
\end{equation}
For the segmentation task, the mean Intersection over Union (mIoU) is used to evaluate the model's performance, computed as follows:
\begin{equation}
    IoU = \frac{TP}{TP+FP+FN}
\end{equation}
\begin{equation}
    mIoU = \frac{1}{C}\sum_{c=1}^{C}IoU_{c}
\end{equation}
To evaluate the performance of the object detection model, we use (mean Average Precision)mAP as our metric. Here we set the threshold as 50\%, which means
IoU over 0.5 will be considered as correct detection. 
\begin{equation}
    mAP = \frac{\sum_{i=1}^{C}AP_{i}}{C}
\end{equation}
\subsection{Evaluation Datasets}
To comprehensively assess the viability of our proposed approach, we conducted experiments across two distinct datasets: Cat-Dog classification\cite{catdogs}, and CityPersons\cite{zhang2017citypersons}. These datasets are representative of key tasks in the computer vision field: classification, segmentation, and object detection, respectively.

The Cat-Dog dataset contains 25,000 images, each labeled as either a cat or a dog.
The CityPersons\cite{zhang2017citypersons} dataset is a subset of Cityscapes, focusing solely on person annotations. It includes 2,975 images for training and 500 images for validation.
The Cambridge driving Labeled Video Database(CamVid)\cite{camVid2009semantic} is the first collection of videos with object class semantic labels, complete with metadata. The dataset contains over 700 images with pixel-level annotation. The annotation of images cover 32 class labels from urban and non-urban driving scenes.



\begin{table}[]
    \centering
    \begin{tabular}{c|ccc}
    \toprule
         \multirow{2}{*}{Methods}&\textbf{Cat-Dog} &\multicolumn{2}{c}{\textbf{CityPersons}} \\
         & (Acc)& (mIoU)& (mAP) \\
    \midrule
    \midrule
       B   &0.927&0.914&0.557 \\
        B+CP  &0.941&0.897& 0.561\\
       B+aug &0.957&0.903& 0.567\\
       B+aug+cp &0.962&0.911& 0.571\\
       B+CACP &0.969&0.929& 0.577\\
        B+CACP+aug  &\textbf{0.974}&\textbf{0.938}& \textbf{0.591}\\
    \bottomrule
    \end{tabular}
    \caption{Results between copy-paste(CP) and context-aware Copy- Paste(CACP) in classification, segmentation and detection tasks.}
    \label{result}
\end{table}

\subsection{Results}
\subsubsection{Across Initialization}
\begin{figure}[htb]
    \centering
    \includegraphics[width=0.45\textwidth]{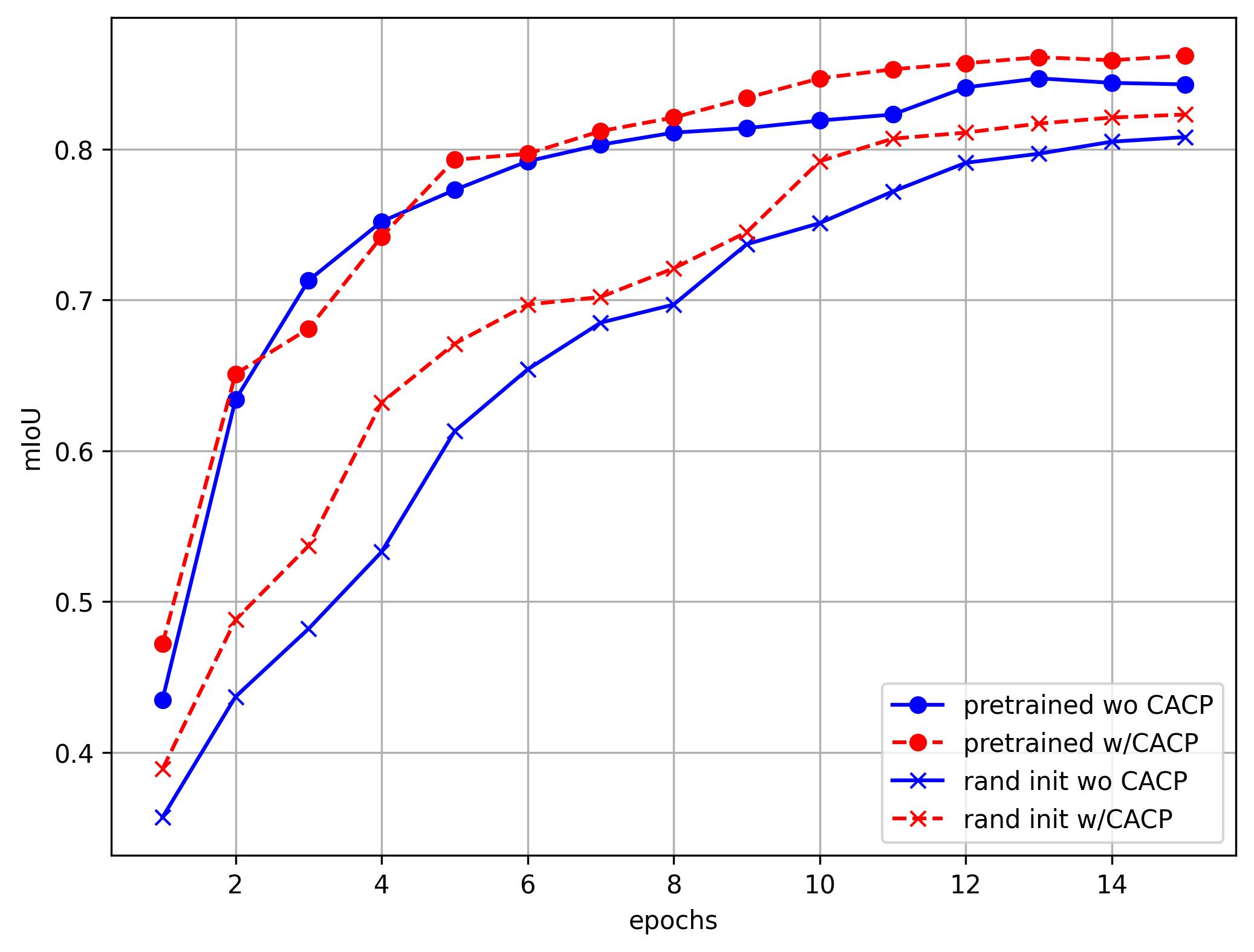}
    \caption{CACP provides gains that are robust to training configurations. We train DeepLabv3 on CamVid for varying number of epochs. The CACP is helpful under with and without pretraining configurations.  }
    \label{initialization}
\end{figure}

To validate the robustness of CACP across different initialization,
we conduct experiments on CamVid based on two different  initialization configurations, namely ImageNet\cite{imagenet2009imagenet} pretrained and normal initialization. As illustrated in \textbf{Fig .}\ref{initialization}, the results with CACP outperform the one without CACP in both configurations.
\subsubsection{Across Tasks} We conduct experiments on different computer vision tasks to validate the usage scenarios of CACP, namely image classification tasks on Cat-Dog, image segmentation and object detection on CityPersons. As illustrated in \textbf{Table }\ref{result}, the items in the column Methods indicate different combinations. \textbf{B} indicates base model. In the classification task this is a Resnet-50 and in segmentation it is DeepLabv3\cite{deeplabv3}, while in object detection it is YOLOv5-s\cite{ultralytics2021yolov5}. \textbf{CP} and \textbf{CACP} indicate random crop-paste and context-aware copy-paste, respectively. \textbf{aug} is a combination of traditional data augmentation techniques, including flip, Color jittering, random noise, which is provided by lib Albumentation\cite{albumentation}. All augmentation techniques have the ability to boost the model. CACP outperforms CP and aug across all three tasks. 
\subsubsection{Across Architectures}
 As illustrated in \textbf{Table} \ref{across architectures}, to validate the effectiveness of our method across different architectures, we conduct experiments on CamVid\cite{camVid} dataset using popular encoder-decoder architectures, namely U-net\cite{unet}, FPN\cite{fpn}, PSPNet\cite{pspnet}, DeepLabv3\cite{deeplabv3}, DeepLabV3+\cite{deeplabv3plus}, and PAN\cite{PAN}. The experiment is conducted with or without our CACP augmentation operation. To observe the effect of different categories, we select six classes: [car,pedestrian,building,tree,sky,road]. It can be noticed that almost all classes results are improved with CACP augmentation (except Sky in DeepLabV3 configuration). 
\subsubsection{Across Partition}
To validate the effect of the number of augmented images, we conduct experiments on CamVid as illustrated in \textbf{Table} \ref{across partition}. \textbf{1/n} indicates 1/n of training images have been augmented using CACP. It is important to note that the total number of training image is fixed, only the ratio of augmented to non-augmented varies.
The result illustrates that the performance increases with the rise of partition from 1/8 to 1/2, the trend is stable in all 4
experiments. The increase is saturated when partition is over 1/2.

\subsubsection{Speed Up Convergence}
\begin{figure}[htb]
    \centering
    \includegraphics[width=0.45\textwidth]{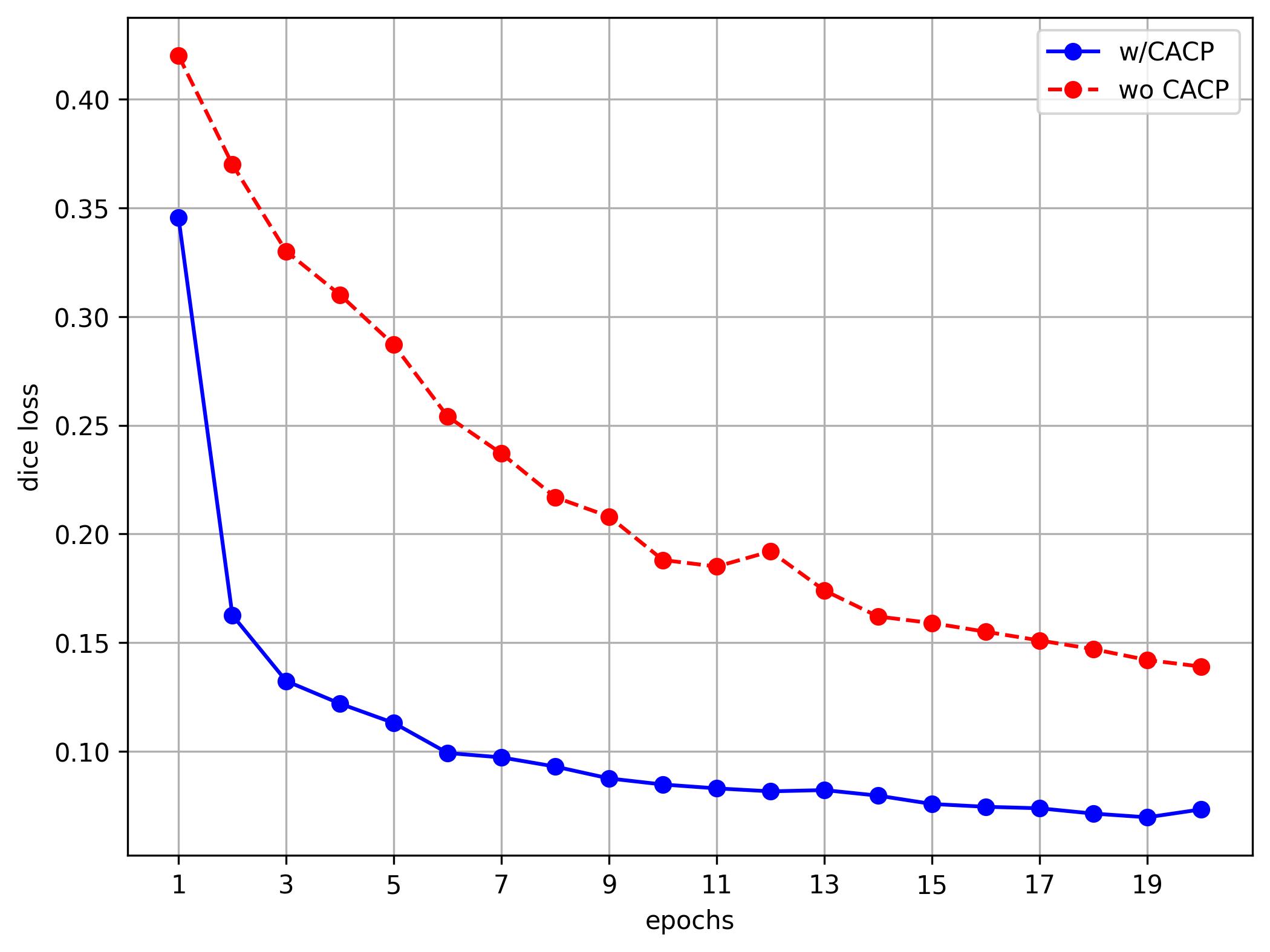}
    \caption{During training, the Dice loss with the CACP configuration converges faster compared to the process without the CACP configuration.}
    \label{loss}
\end{figure}

We have noticed that CACP contributes to speed up the training process. We trained DeepLabv3 on Camvid with 20 epochs. As illustrated in \textbf{Fig.} \ref{loss}, the CACP augmented one converges rapidly compared to the wo-CACP one.
The loss is stable around epoch 15, while the wo-CACP is still not fully converged  after epoch 19.

\subsubsection{Effect of GradCAM Prompt}
To improve the robustness and stability of the output masks of SAM, we propose the GradCAM-guided approach to generate points+bounding box prompts. During our trials, we notice that the number of selected points in GradCAM affects the final segmentation mask. To determine the best point numbers, we conduct the following experiment on CamVid. \textbf{Rand} indicates the point is sampled randomly inside bounding box. 
CAM(n) means n points are extracted in GradCAM map with high value. \textbf{Table} \ref{cam ablation} indicates that extra prompts can improve the accuracy of SAM; CAM-based point prompts are better than random points. The preferred number of points is around 3 to 5.
\begin{table}[htb]
    \centering
    \begin{tabular}{c|cccc}
    \toprule
         \multirow{2}{*}{Methods}&&DeepLabv3&\\
         & 1/8& 1/4& 1/2 &Full\\
    \midrule
    \midrule
       PSPNet   &0.872&0.891&0.887&0.893 \\
        U-NET  &0.851&0.873& 0.879&0.877\\
       PAN &0.862&0.871& 0.877&0.874\\
       DeepLabv3 plus &0.883&0.903& 0.901&0.907\\
    \bottomrule
    \end{tabular}
    \caption{Results between copy-paste(CP) and context-aware Copy- Paste(CACP) in classification, segmentation and detection tasks.}
    \label{across partition}
\end{table}

\begin{table}[htb]
    \centering
    
    \begin{tabular}{ccccc}
    \toprule
        bbox &+rand(1)&+CAM(1)&+CAM(3)&+CAM(5) \\
    \midrule
        0.734&0.841&0.927&0.934&0.933\\
    \bottomrule
    \end{tabular}
    \caption{mIoU across different prompts.The performance improves with the number of CAM-generated point prompts and stabilizes when the number of point prompts exceeds three.}
    \label{cam ablation}
\end{table}

%
\begin{table*}[H]
    \centering
    \begin{tabular}{ccccccc}
    \toprule
         \multirow{2}{*}{Methods}&&&&Camvid &&\\
         &car &pedestrian&building&tree&traffic lane&traffic cone\\
    \midrule
    \midrule
     SAM+B & &&&&&\\
        SAM+B & &&&&&\\
         SAM+B+random(1) && &&&&\\
         SAM+b+CAM(1)&& &&&&\\
          SAM+b+CAM(3)&& &&&&\\
           SAM+b+CAM(5)&& &&&&\\
    \bottomrule
    \end{tabular}
    \caption{Caption}
    \label{tab:my_label}
\end{table*}

\section{Discussion}
In this paper, we propose a context-aware copy-paste (CACP) data augmentation approach, designed as a versatile plug-and-play module for various computer vision tasks, eliminating the need for additional manual annotation. CACP is particularly effective for custom segmentation in semi-supervised learning, offering a time-efficient and scalable solution that allows users to tailor their target gallery to specific task requirements.

Future research directions include the following:
\begin{itemize}
    \item Integration with diffusion models to generate synthetic augmented images with well-crafted prompts, particularly for privacy-sensitive objects, thereby mitigating privacy concerns.
    \item Adaptation to downstream industrial applications, such as obstacle detection and pedestrian detection.
    \item Enhancement through advanced image composition techniques, including object placement, image blending, image harmonization, and shadow generation, to improve the realism and consistency of augmented images
\end{itemize}

\bibliographystyle{ieeenat_fullname}
\bibliography{ref}

\end{document}